\documentclass{article}

 \usepackage[preprint,nonatbib]{neurips_2026}

\usepackage[utf8]{inputenc} 
\usepackage[T1]{fontenc}    
\usepackage{hyperref}       
\usepackage{url}            
\usepackage{booktabs}       
\usepackage{amsfonts}       
\usepackage{nicefrac}       
\usepackage{microtype}      
\usepackage{xcolor}         

\usepackage{hyperref}
\usepackage{url}

\usepackage{amsmath}   
\usepackage{cleveref}
\crefname{equation}{Eq.}{Eqs.}
\Crefname{equation}{Eq.}{Eqs.}
\crefname{section}{Sec.}{Secs.}
\Crefname{section}{Sec}{Secs}
\Crefname{table}{Tab}{Tabs}
\crefname{table}{Tab.}{Tabs.}
\crefname{figure}{Fig.}{Figs.}
\Crefname{figure}{Fig.}{Figs.}
\crefname{algorithm}{Alg.}{Algs.}
\Crefname{algorithm}{Alg.}{Algs.}
\crefname{appendix}{App.}{Apps.}
\Crefname{appendix}{App.}{Apps.}

\usepackage{graphicx}
\usepackage{subcaption}
\usepackage{caption}
\usepackage{adjustbox}
\usepackage{wrapfig}
\usepackage{algorithm}
\usepackage[noend]{algpseudocode}
\usepackage{dsfont}
\usepackage{booktabs}
\usepackage{multirow}
\usepackage{multicol}
\usepackage{colortbl}
\usepackage{xcolor}
\usepackage{mathtools}
\usepackage{makecell}

\usepackage{array}
\usepackage{tabularray}
\usepackage{tabularx}
\usepackage{enumitem}
\usepackage{listings}
\lstdefinelanguage{prompt}{
    basicstyle=\linespread{1.1}\ttfamily\scriptsize, 
    breaklines=true,                 
    frame=single,                    
    rulecolor=\color{gray!30},       
    backgroundcolor=\color{gray!5},  
    keywordstyle=\bfseries\color{blue},
    commentstyle=\itshape\color{green!50!black},
    stringstyle=\color{red},
    extendedchars=true,
    showstringspaces=false,
    captionpos=b,                    
    aboveskip=1em,
    belowskip=1em
}

\def\makebold#1{\expandafter\def\csname m#1\endcsname{\mathbf{#1}}}
\makebold{w}
\makebold{x}
\makebold{y}
\makebold{g}
\makebold{z}

\title{MobileKernelBench: Can LLMs Write Efficient Kernels for Mobile Devices?}

\author{
Xingze Zou $^{1}$ \thanks{Equal Contribution} \quad
Jing Wang $^{1}$ \footnotemark[1] \quad
Yuhua Zheng$^{1}$ \quad
Xueyi Chen$^{2}$ \quad
Haolei Bai$^{2}$ \quad
Lingcheng Kong$^{3}$ \quad
\\
\vspace{-2mm}
\textbf{Syed A.R. Abu-Bakar}$^{4}$ \quad
\textbf{Zhaode Wang}$^{5}$ \quad
\textbf{Chengfei Lv}$^{5}$ \quad
\textbf{Haoji Hu}$^{1}$ \thanks{Corresponding Author}\quad
\textbf{Huan Wang}$^{2}$  \footnotemark[2]\\
\\
$^{1}$Zhejiang University  $^{2}$Westlake University $^{3}$HKUST $^{4}$Universiti Teknologi Malaysia $^{5}$Alibaba \\
\vspace{-2mm}
\\
\texttt{\{zeezou, j\_wang, YuhuaZheng, haoji\_hu\}@zju.edu.cn} \\
\texttt{\{chenxueyi, wanghuan\}@westlake.edu.cn, Baih0011@e.ntu.edu.sg} \\
\texttt{LingchengKong05@outlook.com, e-syed@utm.my} \\
\texttt{\{zhaode.wzd, chengfei.lcf\}@alibaba-inc.com} 
}

\begin{document}
\maketitle

\vspace{-4mm}
\begin{abstract}
\vspace{-1mm}
Large language models (LLMs) have demonstrated remarkable capabilities in code generation, yet their potential for generating kernels specifically for mobile devices remains largely unexplored. 
In this work, we extend the scope of automated kernel generation to the mobile domain to investigate the central question: \textbf{Can LLMs write efficient kernels for mobile devices?} %
To enable systematic investigation, we introduce \textbf{MobileKernelBench}, a comprehensive evaluation framework comprising a benchmark prioritizing operator diversity and cross-framework interoperability, coupled with an automated pipeline that bridges the host-device gap for on-device verification. %
Leveraging this framework, we conduct extensive evaluation on the CPU backend of Mobile Neural Network (MNN), revealing that current LLMs struggle with the engineering complexity and data scarcity inherent to mobile frameworks; standard models and even fine-tuned variants exhibit high compilation failure rates (over 54\%) and negligible performance gains due to hallucinations and a lack of domain-specific grounding. %
To overcome these limitations, we propose the \underline{Mo}bile \underline{K}ernel \underline{A}gent (\textbf{MoKA}), a multi-agent system equipped with repository-aware reasoning and a plan-and-execute paradigm. %
Validated on MobileKernelBench, MoKA achieves state-of-the-art performance, boosting compilation success to 93.7\% and enabling 27.4\% of generated kernels to deliver measurable speedups over native libraries.  Our dataset is available on \href{https://zeezou-isee.github.io/Mobilekernelbench/}{Mobilekernelbench}.
\vspace{-3mm}
\end{abstract}

\section{Introduction}

\begin{figure*}[t]
  \centering
  \includegraphics[width=0.85\linewidth]{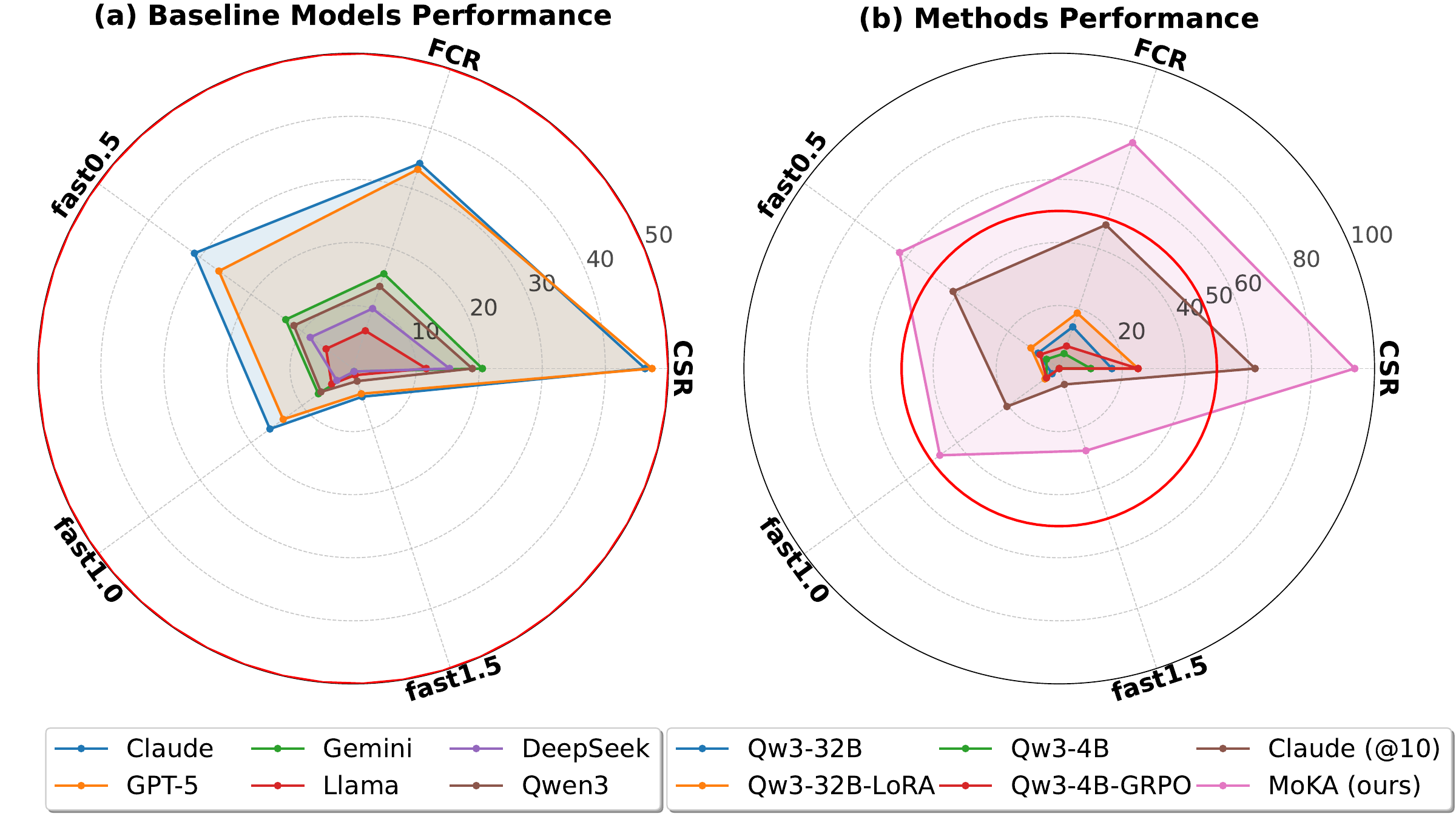}
  \caption{\textbf{ Performance evaluation on MobileKernelBench across three metrics:} compilation success rate (CSR), functional correctness rate (FCR), and performance speedup (fast$_p$) \cite{ouyang2025kernelbench}. \textbf{(a) Baseline LLM performance:} We benchmark prevalent open- and closed-source LLMs, revealing significant shortcomings in their ability to generate functional and efficient mobile kernels. \textbf{(b) Method comparison}: We compare our proposed MoKA against common training methods, including LoRA and GRPO. The red circle (marked at 50\%) corresponds to the outer limit of plot (a), highlighting that MoKA achieves substantial improvements, surpassing the performance ceiling of both baseline models and naive fine-tuning approaches.}
\vspace{-3mm}

  \label{fig:total_comparison}
\end{figure*}
\label{sec:intro}
Large language models (LLMs) have achieved remarkable success in the field of code generation, with performance further enhanced by specialized methodologies such as domain-adaptive fine-tuning and agentic reasoning frameworks \cite{openai_gpt5_2025, anthropic_claude_sonnet45, deepmind_gemini, yang2025qwen3, team2025kimi, zeng2025glm}. %
Building on this foundation, a series of recent studies have begun to systematically evaluate the application of LLMs in synthesizing performance-optimized CUDA kernels \cite{ouyang2025kernelbench, wen2025multikernelbench, li2025tritonbench}. %
These works demonstrate that LLMs can exhibit a significant degree of hardware awareness, enabling them to write high-performance kernels tailored for server-grade GPUs. %
With the rapid surge in demand for mobile AI applications, on-device inference has attracted increasing attention for its data safety, low inference latency, and personalized service. %
However, deploying deep learning models on mobile devices also demands substantial effort in kernel development, constituting a highly challenging engineering barrier that has not yet been investigated. %
In this work, we extend the scope of automated kernel generation to the mobile domain, presenting a preliminary investigation into the following question: \textit{Can LLMs write efficient kernels for mobile devices?}

We first investigate and compare the differences between server-side and mobile-side kernel development in~\cref{tab:server_side_vs_mobile}. %
Mobile kernel development functions as the critical bridge for migrating models from training environments to resource-constrained edge devices, characterized by three distinct features: %
(1) \textbf{Compatibility priority}: the primary objective is to ensure broad support, necessitating the implementation of a vast spectrum of operators to cover diverse training frameworks; %
(2) \textbf{Engineering complexity}: the development process, however, is severely hampered by the fragmentation of the mobile ecosystem, where developers must navigate a myriad of heterogeneous backends and architectures; %
and (3) \textbf{Data scarcity:} the nascent nature of the mobile inference landscape results in a lack of high-quality reference implementations, creating a data-poor environment that poses significant generalization challenges for LLMs. %
Consequently, existing research is ill-suited for this specific domain: current benchmarks prioritize the algorithmic complexity of kernels rather than the operator diversity required for edge compatibility, while the tight coupling between kernels and the corresponding framework has precluded the establishment of systematic pipelines for evaluating LLM-generated kernels directly on mobile devices.

To address these limitations, we introduce \textbf{MobileKernelBench}, as illustrated in \cref{fig:overview}, a comprehensive system comprising a dedicated benchmark and an automated evaluation pipeline tailored for mobile frameworks. %
Diverging from the difficulty-tiered classification of prior works \cite{ouyang2025kernelbench,wen2025multikernelbench}, our benchmark prioritizes operator diversity by curating 190 tasks across 12 categories of 95 primitive operators to facilitate wider model migration. %
Simultaneously, we ensure cross-framework interoperability via standardized PyTorch and Open Neural Network Exchange (ONNX) test data, acting as a universal bridge to mitigate inconsistent framework support and tackle ecosystem fragmentation. %
Complementing the benchmark, we introduce an automated evaluation pipeline to bridge the separation between host-side development and device-side testing inherent to mobile deployment. %
By automating the entire lifecycle spanning registration, cross-compilation, and on-device verification, it not only satisfies the concerns mentioned in previous work \cite{ouyang2025kernelbench} but also streamlines the desktop-to-mobile workflow to capture granular debugging and performance data akin to a real-world developer's environment.

Building on the proposed evaluation pipeline design, we establish a concrete evaluation environment using Mobile Neural Network (MNN) \cite{jiang2020mnn} framework with a CPU backend. %
We first conduct a comprehensive evaluation of state-of-the-art (SOTA) LLMs on MobileKernelBench. Our experimental results show that, due to limited framework-specific knowledge and insufficient optimization capability, LLM-generated kernels achieve performance parity with native framework implementations in at most 16.3\% of benchmark cases, illustrated in \cref{fig:total_comparison} (a). Strikingly, more than 54.7\% of the generated kernels fail at the compilation stage, primarily driven by hallucinated APIs or invalid framework usage, revealing a critical lack of grounding in framework-specific logic. Notably, only a small fraction of kernels achieve measurable speedups over the baseline. %
Subsequently, we apply standard training strategies, including LoRA \cite{hu2022lora} and GRPO \cite{shao2024deepseekmath}, yet observe negligible improvements. We attribute these persistent failures to the scarcity of high-quality data within specific mobile inference frameworks. %
This data poverty creates a severe deficit in domain-specific knowledge, spanning framework specifications, optimization heuristics, and functional definitions, preventing LLMs from mastering this highly specialized domain.%

To further explore the potential of LLMs in mobile kernel generation and address their identified limitations, we propose the \underline{Mo}bile \underline{K}ernel \underline{A}gent (\textbf{MoKA}), a specialized multi-agent system tailored for mobile kernel development. %
The MoKA follows a multi-round plan-and-execute paradigm, consisting of a code agent responsible for operator generation and two planning agents that respectively formulate execution strategies for compilation, functional correctness verification, and performance optimization. %
These agents are equipped with repository-aware and information parsing tools, enabling them to access and reason over realistic deployment signals. %
As illustrated in \cref{fig:total_comparison} (b), when evaluated on MobileKernelBench, the MoKA substantially outperforms the baseline and establishes SOTA performance among all tested LLMs, with $93.7\%$ of kernels achieving successful compilation and $27.4\%$ delivering measurable performance speedups. This initiative advances the frontier of mobile kernel generation, offering significant insights into enhancing LLM capabilities within such highly specialized domains.

In summary, our contributions are as follows:
\begin{itemize}
    \item We present the first systematic study extending automated kernel generation to the mobile domain, identifying three distinguishing features of mobile development and revealing the fundamental gap between server-side and mobile-side development.
    \item We introduce \textbf{MobileKernelBench}, a comprehensive evaluation system comprising a benchmark and an automated pipeline, facilitating systematic mobile kernel evaluation and granular information capture.
    \item We propose \textbf{MoKA}, a multi-agent system designed to autonomously navigate the complexities of API usage and performance optimization in data-scarce environments.
    \item We conduct extensive experiments on MobileKernelBench, revealing that while standard LLMs suffer from severe hallucinations and compilation failures, MoKA effectively overcomes these barriers, achieving SOTA performance over other LLMs.
\end{itemize}

\section{Related Work}
\subsection{LLM for Kernel Generation}
LLMs have shown proficiency in general code generation \cite{guo2024deepseek,hui2024qwen2,openai_gpt5_2025,anthropic_claude_sonnet45}, and recent works have focused on leveraging them for high-performance computing \cite{waghjale2024ecco,huang2024effibench}. 
To systematically assess these capabilities, benchmarks such as KernelBench \cite{ouyang2025kernelbench}, MultiKernelBench \cite{wen2025multikernelbench}, and TritonBench \cite{li2025tritonbench} have been developed, evaluating both correctness and efficiency across diverse platforms and paradigms. 
Recognizing the limitations of direct prompting, subsequent studies have integrated iterative feedback loops to refine kernel quality. 
One line of research, including Kevin \cite{baronio2025kevin}, AutoTriton \cite{li2025autotriton}, CUDA-L1 \cite{li2025cuda}, and CUDA-L2 \cite{su2025cuda}, uses reinforcement learning (RL) or supervised fine-tuning (SFT) to optimize kernel generation. 
Alternatively, agentic frameworks such as Astra \cite{wei2025astra}, EvoEngineer \cite{guo2025evoengineer}, and CudaForge \cite{zhang2025cudaforge} employ collaborative role-playing or Coder-Judge architectures to ground reasoning in hardware specifications. 
However, these efforts predominantly focus on server-grade GPUs and CUDA, leaving the distinct constraints and fragmented ecosystems of mobile and edge computing largely unexplored.

\subsection{Mobile Inference Engine}
The Open Neural Network Exchange (ONNX) \cite{onnx} serves as the de facto standard for model interoperability. 
By defining a unified intermediate representation and operator schema, ONNX decouples model architecture from specific training environments like PyTorch \cite{paszke2019pytorch} and TensorFlow \cite{abadi2016tensorflow}. 
This standardization is particularly critical for benchmarking, as it establishes a universal functional definition for operators, ensuring that evaluations focus on implementation capability rather than framework discrepancies.
For on-device execution, bridging the gap between abstract representation and efficiency remains challenging. 
While compiler-based stacks like TVM \cite{chen2018tvm} offer automation, library-based inference engines such as MNN \cite{jiang2020mnn} and NCNN \cite{Ni_ncnn_2017} are often preferred in industry for their lightweight integration. 
Achieving high performance in these libraries requires meticulous hardware-aware optimizations, including dynamic algorithmic selection and specialized memory layouts. 
Consequently, manually implementing operators involves managing complex low-level details like data packing and register allocation, creating a significant barrier that motivates the need for automated, expert-level code generation approaches.


\newcolumntype{Y}{>{\centering\arraybackslash}X} 

\begin{table}[t]
\centering
\caption{\textbf{Comparison between server-side and mobile-side computing.} Unlike the server side, which utilizes unified CUDA-based backends on high-performance GPUs, the mobile side faces a fragmented landscape with diverse compute backends primarily optimized for inference tasks.}

\label{tab:server_side_vs_mobile}

\def\TableWidth{\textwidth}
\renewcommand{\arraystretch}{1.1} 
\small

\begin{tabularx}{\TableWidth}{l Y Y} 
\toprule
\textbf{Dimension} & \textbf{Server Side} & \textbf{Mobile Side} \\
\midrule
Hardware  & CPU, GPU, TPU, NPU (\textit{ASIC}) & CPU, GPU, NPU (\textit{SoC}) \\
\rowcolor{gray!15} Backend & CUDA, TensorRT, ROCm & CPU, OpenCL, Vulkan, CoreML, Metal \\
Language & CUDA, Triton, PTX & C++, Assembly \\
\rowcolor{gray!15} Feature & Homogeneous & Heterogeneous \& Fragmented \\
Frameworks & PyTorch, TensorFlow, JAX & MNN, NCNN, TFLite, CANN \\
\rowcolor{gray!15} Workload & Training \& Inference & Inference \\
Resource & Unlimited & Limited (Power/Mem/Band) \\
\bottomrule
\end{tabularx}

\end{table}

\section{MobileKernelBench}
To bridge the gap between existing server-centric benchmarks and the specialized requirements of on-device inference, we introduce \textbf{MobileKernelBench}, the first benchmark dedicated to cross-framework kernel implementation for mobile deployment. As detailed in \cref{tab:dataset_comparison}, our benchmark differs from previous work by focusing on the ONNX standard and incorporating a mobile kernel development evaluation pipeline, which verifies the full lifecycle of on-device integration.

\subsection{Data Curation}

\newcolumntype{Y}{>{\centering\arraybackslash}X} 
\begin{table}[t]
\vspace{-3mm}
\centering
\caption{\textbf{Comparison between KernelBench and MobileKernelBench. }Compared to KernelBench, MobileKernelBench emphasizes operator diversity, covering the majority of common operator types in ONNX Opset 20. The (PyTorch, ONNX) pairs format facilitates kernel evaluation across different frameworks, targeting both framework adaptation and optimization on mobile devices.}

\label{tab:dataset_comparison}

\def\TableWidth{\textwidth}
\renewcommand{\arraystretch}{1.1} 
\small

\begin{tabularx}{\TableWidth}{l Y Y} 
\toprule
\textbf{Feature} & \textbf{KernelBench} & \textbf{MobileKernelBench (Ours)} \\
\midrule
\rowcolor{gray!15} Diversity & 43 (Level 1) & 95  \\
Tasks & 250  & 190  \\
\rowcolor{gray!15} Format & PyTorch Models & (PyTorch, ONNX) Pairs \\
Domain & GPUs  & Mobile SoCs (CPU, Adreno, etc.) \\
\rowcolor{gray!15} Target & Optimization ONLY & Framework Adaptation \& Optimization. \\
\bottomrule
\vspace{-8mm}
\end{tabularx}
\end{table}

\begin{figure}
\vspace{-4mm}
  \centering
    \centering
    \includegraphics[width=\linewidth]{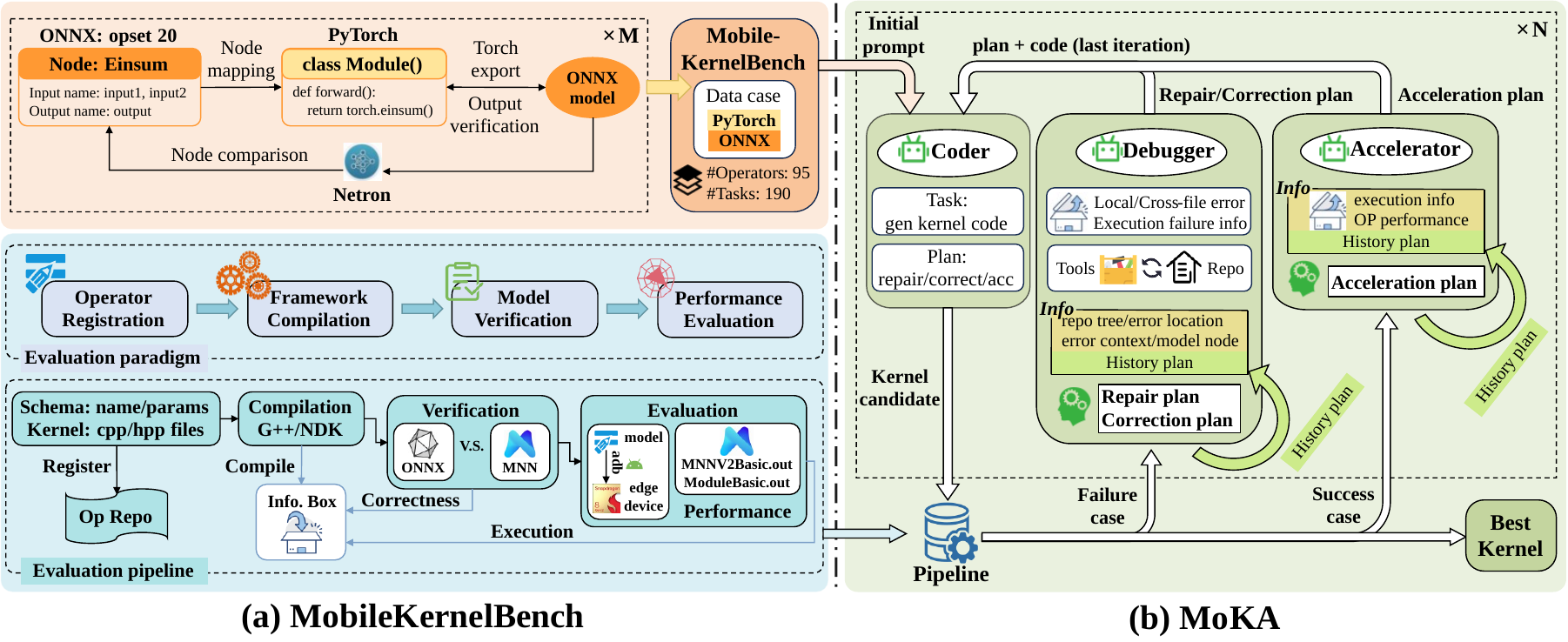}
    \captionof{figure}{\textbf{Overview of our proposed framework.}
    The system consists of two core components: 
    \textbf{(a) MobileKernelBench}, which establishes the evaluation environment by integrating a target-driven data curation process with an automated, hardware-in-the-loop evaluation pipeline; and 
    \textbf{(b) MoKA}, a multi-role agentic system where Coder, Debugger, and Accelerator agents collaborate to iteratively generate and refine kernels based on feedback from the benchmark.}
    \vspace{-7mm}
    \label{fig:overview}
\end{figure}

Motivated by the mobile ecosystem's reliance on ONNX as the standard interchange format, our benchmark prioritizes comprehensive coverage of the ONNX operator set (Opset). To ensure compatibility with existing kernel generation benchmarks and standard LLM evaluation protocols, we encapsulate these operators within a PyTorch-based task format. To construct a usable, diverse, and robust dataset, we execute a systematic curation pipeline consisting of three stages.

\textbf{Standard-aligned operator collection.}
We ground our benchmark in the ONNX Opset version 20 to ensure compatibility with modern inference engines. From the full ONNX operator catalog, we select operators according to two principles: diversity and generality. To ensure diversity, we include operators from a wide range of categories to reflect the heterogeneous computation patterns in neural network inference. To ensure generality, we prioritize operators that are widely used in practice, including those commonly appearing in canonical neural network architectures and those already supported by mainstream mobile inference frameworks such as MNN \cite{jiang2020mnn} and NCNN \cite{Ni_ncnn_2017}. Following these criteria, we select 95 representative operators that capture the core workloads of on-device inference while keeping the benchmark compact.

\textbf{Multi-dimensional task expansion.}
A single operator implementation is often insufficient to capture the complexity of varying attributes and input configurations. To rigorously evaluate the generalization capability of LLMs, we expand the 95 selected operators into 190 specific tasks through a task expansion strategy. Instead of relying on default parameters, we systematically mutate the PyTorch model definitions to cover diverse code paths and edge cases. Specifically:
\begin{itemize}[topsep=0pt, itemsep=1pt, parsep=0pt]
    \item \textbf{Attribute variation:} We modify key attributes that directly affect control-flow behavior. For example, in reduction operators such as \texttt{ArgMax} and \texttt{ReduceSum}, we vary the \texttt{keep\_dims} parameter and the reduction axis to evaluate the model's ability to handle output shape inference and loop reduction logic.
    \item \textbf{Dimensionality transformation:} We extend input tensors from standard 2D matrices to lower or higher-dimensional forms. For matrix multiplication operators including \texttt{MatMul} and \texttt{Gemm}, this requires the generated kernel to correctly handle broadcasting semantics and batch stride arithmetic, significantly increasing the implementation difficulty.
\end{itemize}

\textbf{Target-driven construction strategy.}
To ensure the expanded tasks are valid and align with deployment standards, we employ a target-driven construction strategy. We actively construct paired (PyTorch, ONNX) instances as illustrated in \cref{fig:overview}(a) and filter them through two criteria:

\begin{itemize}[topsep=0pt, itemsep=1pt, parsep=0pt]
    \item \textbf{Output fidelity:} The implemented PyTorch module must yield results identical to the target ONNX operator. We verify this by cross-referencing the execution of the PyTorch code in eager mode against the ONNXRuntime \footnote{https://github.com/microsoft/onnxruntime} execution of the exported model.
    \item \textbf{Topology alignment:} The PyTorch implementation must be converted to the precise target ONNX operator node rather than a composite subgraph. We validate this by inspecting the exported graph topology with Netron \footnote{https://github.com/lutzroeder/netron}, iteratively refining the source code until the graph structure aligns with the canonical ONNX definition.
\end{itemize}

The resulting benchmark comprises 190 tasks covering 95 distinct operators, taxonomized into 12 categories in \cref{tab:MobileKernelBench}. This set spans diverse computational patterns, ranging from memory-bound element-wise operations to compute-bound contractions like convolutions.

\begin{table}[t]
\centering
\vspace{-2mm}
\caption{\textbf{Overview of MobileKernelBench.} This benchmark comprises 190 tasks derived from 95 primitive operators. These operators are classified into 12 categories, encompassing common operators found in the ONNX ecosystem. A primitive operator may yield multiple distinct tasks based on differences in input shapes or parameter settings.}
\label{tab:MobileKernelBench}

\def\TableWidth{\textwidth}
\renewcommand{\arraystretch}{1.1}
\small
\begin{tabularx}{\TableWidth}{
    >{\centering\arraybackslash}p{2cm}
    >{\centering\arraybackslash}X
    >{\centering\arraybackslash}p{2cm}
    >{\centering\arraybackslash}p{2cm}
}
\toprule
\textbf{Categories} & \textbf{Representatives} & \textbf{\#Operators} & \textbf{\#Tasks} \\
\midrule
Unary         & exp, sign, ceil, floor          & 11       & 11 \\
Binary        & add, div, mod                    & 7        & 8 \\
Trigonometry  & sin, acos, atanh, tan         & 12       & 12 \\
Activation    & hardsigmoid, softmax, celu, relu         & 6        & 12 \\
Normalization & batchnorm, layernorm, instancenorm        & 3        & 12 \\
Pooling       & maxpool, averagepool         & 4        & 15 \\
Convolution   & conv2d, convtranspose,        & 2        & 21 \\
Matrix        & einsum, gemm, matmul, det        & 4        & 19 \\
Reduction     & reducemin, reduceprod, reducesum        & 10       & 35 \\
Tensor        & reshape, concat, topk, tile, slice       & 19       & 28 \\
Logic         & and, bitwisexor, equal       & 11       & 13 \\
Others        & RNN, LSTM, STFT, RoiAlign             & 4        & 4  \\
\midrule
\textbf{Total}&  & \textbf{95} & \textbf{190} \\
\bottomrule
\end{tabularx}
\vspace{-4mm}
\end{table}

\subsection{Evaluation Pipeline}

Evaluating mobile kernels presents a unique challenge: code generation occurs on host machines, while verification must be executed on resource-constrained edge devices. To address this, we propose a generic evaluation protocol for mobile kernels, which is instantiated through an automated, end-to-end pipeline as shown in \cref{fig:overview}(a). For more detailed configurations, see \cref{appdix:pipeline}

\textbf{Evaluation paradigm.} We design a standardized four-stage protocol comprising \textit{Operator Registration}, \textit{Framework Compilation}, \textit{Model Verification}, and \textit{Performance Evaluation}. This paradigm explicitly decouples operator implementation from runtime execution. By formalizing this separation, we ensure the evaluation remains reproducible and accurately simulates the deployment workflow of real-world mobile applications.

\textbf{Pipeline instantiation on MNN.} We instantiate this paradigm within the MNN framework, automating the operator lifecycle through four specific stages:
(1) \textbf{Automated registration.} Upon code generation, the pipeline parses the output and executes an injection strategy. It locates the target source path and hot-swaps the existing implementation with the generated code, ensuring seamless registration into the global operator factory without manual intervention.
(2) \textbf{Framework compilation.} The pipeline invokes the CMake build system to compile the modified MNN library. This stage acts as a strict filter for syntactic correctness and API validity, where errors trigger immediate failure termination.
(3) \textbf{Functional verification.} Successfully compiled operators undergo differential testing. We establish a ground truth baseline using the reference ONNX model with randomized inputs. The pipeline converts the ONNX model to MNN format and compares the MNN inference output against the baseline, enforcing a strict numerical tolerance for validation.
(4) \textbf{On-device performance benchmarking.} We employ a cross-compilation and remote execution workflow for real-world efficiency evaluation. The modified source is cross-compiled via the Android NDK, and the resulting ARM64 binaries are deployed to the target device via a bridge interface. To ensure statistically robust measurements, the benchmarking executable follows a strict warm-up and multi-iteration protocol to mitigate system noise.

\section{MoKA}
We propose the \underline{Mo}bile \underline{K}ernel \underline{A}gent (MoKA), a specialized multi-agent framework designed to automate the lifecycle of operator implementation, debugging, and optimization for on-device inference engines. As shown in \cref{fig:overview} (b), MoKA follows an iterative \textit{plan-and-execute} paradigm. In each iteration, the agent generates an operator candidate that probes the deployment environment, while the evaluation pipeline returns structured feedback including compilation status, correctness, and performance metrics. This feedback loop drives the agents to refine their planning strategies, enabling progressive convergence toward high-quality, deployment-ready operators.

\subsection{Agent Collaboration Design}
The MoKA architecture decomposes the implementation task into three specialized roles that collaborate via a shared history memory.

\textbf{Coder.} The Coder serves as the sole actuator of the system, responsible for synthesizing C++ source code. In the initial iteration, it generates a draft implementation based on the task description. In subsequent steps, it acts as an executor that translates high-level strategies provided by planning agents into concrete code modifications, ensuring adherence to the specified repair or optimization logic. \textbf{Debugger.} Activated upon pipeline failures, the Debugger handles two distinct error modes. For build failures, it initiates a \textit{Compilation Diagnosis (Repair Plan)} by utilizing repository-aware tools to interpret compiler diagnostics and retrieve cross-file context, thereby formulating plans to fix syntactic or dependency errors. For incorrect outputs, it initiates a \textit{Functional Correction (Correction Plan)} by employing model parsing tools to align the implementation with the ONNX definition, identifying semantic discrepancies to generate functionality-preserving fixes. \textbf{Accelerator.} Once an operator passes functional verification, the Accelerator pushes performance boundaries. It leverages a performance parser to extract fine-grained execution metrics such as backend selection and threading efficiency. Combined with historical data, it proposes an \textit{Acceleration Plan} to optimize memory access patterns or computational logic without compromising correctness.

\textbf{Reflective memory.} Crucially, both the Debugger and Accelerator utilize a shared history mechanism. By maintaining a trace of past strategies and their outcomes, the agents perform self-reflection to avoid repetitive mistakes and prune ineffective optimization paths.

\subsection{Agentic Toolset}
To ground agent reasoning within the rigid constraints of the deployment environment, we equip them with two categories of domain-specific tools.

\textbf{Repository-aware tools.} To address the lack of structural awareness in generic LLMs, we implement a \textit{Repository Tree Builder} that efficiently constructs a hierarchical view of the codebase, facilitating the resolution of dependency errors. Additionally, an \textit{Error Extractor} powered by \textit{tree-sitter-cpp} cite{tree-sitter2025} parses compiler logs to pinpoint exact error locations and extract surrounding context. These tools enable the Debugger to effectively distinguish between local syntax errors and complex cross-file inconsistencies.
\textbf{Information parsing tools.} To bridge static graphs and dynamic execution, a \textit{Model Parser} serializes both reference ONNX nodes and converted MNN operators into a unified structured representation, providing evidence for semantic mismatches. Furthermore, a \textit{Performance Parser} processes raw profiling logs (e.g., backend usage, execution time) into structured indicators. This assists the Accelerator in identifying bottlenecks, such as suboptimal data layouts or thread contention, to formulate backend-consistent optimization strategies.
\section{Experiments}
In this section, we conduct a comprehensive evaluation on MobileKernelBench to assess the performance of LLMs in mobile kernel generation. In \cref{sec:baseline_result}, we first evaluate leading open-source and closed-source models accessed directly via cloud service APIs. Furthermore, to investigate the efficacy of standard fine-tuning strategies for this task, we implement classic LoRA fine-tuning and GRPO training.  Finally, we evaluate the performance of our proposed MoKA in \cref{sec:MoKA_result}.
\subsection{Experiment Setups}
\textbf{Evaluation environment.} We utilize MNN (v3.2.2 \footnote{https://github.com/alibaba/MNN/releases/tag/3.2.2}) as the foundational inference engine, specifically targeting its CPU backend for operator registration and execution (see \cref{appdix:MNN_OP} for registration details). All evaluations are conducted on a Xiaomi 13 smartphone powered by the Qualcomm Snapdragon 8 Gen 2 platform \cite{qualcomm_snapdragon8gen2_productbrief_2022}, a representative high-end mobile SoC equipped with an 8-core Kryo CPU and LPDDR5X memory.

\textbf{Evaluation metrics.} We assess performance on MobileKernelBench using three metrics: (1) \textbf{Compilation success rate (CSR)}, the percentage of tasks where generated operators successfully pass the build process. (2) \textbf{Functional correctness rate (FCR)}, the proportion of tasks that pass functional verification against the ONNX baseline. (3) \textbf{fast$_p$}~\cite{ouyang2025kernelbench}, this metric quantifies the percentage of tasks achieving a speedup greater than a threshold p relative to the native MNN implementation. For methods that involve iterative refinement or generate multiple candidates, we report metrics based on the best-performing operator for each task.

\textbf{LoRA training setups.} We fine-tune Qwen-32B via LLaMA-Factory \cite{zheng2024llamafactory} using LoRA (rank 64) and ZeRO stage-3 \cite{rajbhandari2020zero} on 8 A100 GPUs. The model is trained for 2 epochs with a batch size of 1 and gradient accumulation steps of 2. The training dataset follows the Alpaca format, pairing GPT-5 generated descriptions with MNN implementations of 74 ONNX operators. A subset of 20 operators and 36 \textit{MobileKernelBench} tasks is held out for testing.

\textbf{GRPO training setups.} We explore RL using GRPO \cite{shao2024deepseekmath} implemented via the verl \cite{sheng2024hybridflow} framework, with Qwen3-4B-Instruct-2507 as the policy model. Our \textit{MobileKernelBench} is partitioned into a training set of 150 samples and a test set of 40 samples using stratified sampling to ensure a balanced representation of operator categories. Based on the reward scheme from Kevin \cite{baronio2025kevin}, we add an intermediate compilation reward to avoid extremely sparse rewards in early training stages.
Training runs on two A100 GPUs for 40 steps, using a global batch size of 30, a learning rate of 1e-6, a group size of 5, and a context length of 8192 tokens.
\vspace{-2mm}

\begin{figure}[t]
  \centering
  \begin{minipage}[t]{0.48\linewidth}
    \vspace{0pt}
    \centering
    \small
    \renewcommand{\arraystretch}{1.1}
    \setlength{\tabcolsep}{3.5pt}
    \captionof{table}{\textbf{Performance comparison of SOTA LLMs on MobileKernelBench.}
    The best and second-best results are highlighted in \textbf{bold} and \underline{underlined} respectively. Some model names are abbreviated for layout purposes, with full identifiers provided in \cref{sec:baseline_result}.}
    \label{tab:api_results}
    \begin{tabular}{l c c c c c}
      \toprule
      \multirow{2}{*}{\textbf{Baseline}} & \textbf{CSR} & \textbf{FCR} & \multicolumn{3}{c}{\textbf{fast$_p$}} \\
      \cmidrule(lr){4-6}
       & (\%) & (\%) & \textbf{0.5} & \textbf{1.0} & \textbf{1.5} \\
      \midrule
      Claude-Sonnet-4.5 & \underline{46.3} & \textbf{34.2} & \textbf{31.1} & \textbf{16.3} & \textbf{4.7} \\
      GPT-5 & \textbf{47.4} & \underline{33.2} & \underline{26.3} & \underline{13.7} & \underline{4.2} \\
      Gemini-2.5-Flash & 20.5 & 15.8 & 13.2 & 6.8 & 0.5 \\
      Llama-3.1-405B & 11.6 & 6.3 & 5.3 & 4.2 & 1.1 \\
      DeepSeek-R1 & 15.3 & 10.0 & 8.4 & 3.2 & 0.5 \\
      Qwen3-235B & 18.9 & 13.7 & 11.6 & 6.3 & 2.1 \\
      \bottomrule
    \end{tabular}
  \end{minipage}
  \hfill
  \begin{minipage}[t]{0.48\linewidth}
    \vspace{0pt}
    \centering
    \includegraphics[width=\linewidth]{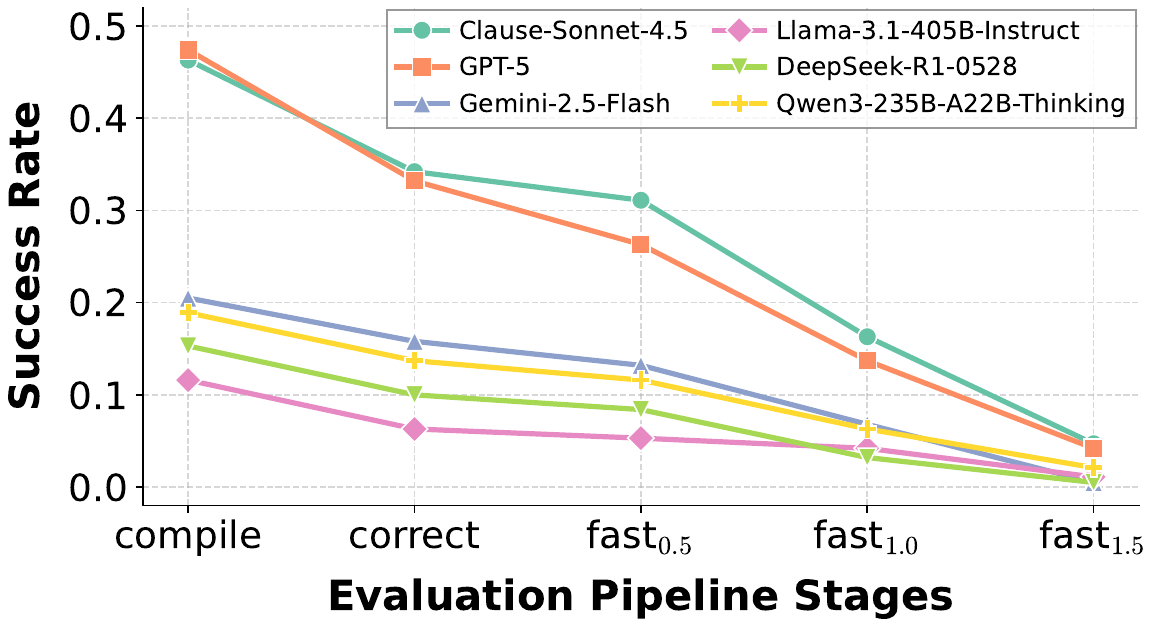}
    \captionof{figure}{\textbf{Success rate degradation across evaluation stages.} 
    The plot illustrates the performance drop of each model as the evaluation criteria become stricter, from compilation to functional correctness and varying levels of performance optimization.
    }
    \label{fig:api_results}
  \end{minipage}
\vspace{-4mm}
\end{figure}

\subsection{Baseline Evaluation}
\label{sec:baseline_result}
We benchmark six prevalent LLMs, spanning proprietary leaders (OpenAI GPT-5 \cite{openai_gpt5_2025}, Anthropic Claude-Sonnet-4.5 \cite{anthropic_claude_sonnet45}, Google Gemini-2.5-Flash \cite{deepmind_gemini}) and open-source frontiers (LLaMA-3.1-405B-Instruct \cite{dubey2024llama}, DeepSeek-R1-0528 \cite{guo2024deepseek}, Qwen3-235B-A22B-Thinking-2507 \cite{yang2025qwen3}). All models are evaluated using a standardized initial prompt (see \cref{appdix:initial_prompt}) under default settings via API endpoints.

The quantitative results presented in \cref{tab:api_results} and \cref{fig:api_results} demonstrate a substantial gap between code generation capabilities and deployment readiness across all evaluated models. Although leading proprietary models, such as Claude-Sonnet-4.5 and GPT-5, achieve compilation success rates of approximately 47\%, their performance declines significantly when strict functional verification criteria are applied. Furthermore, open-source models demonstrate considerably lower proficiency, with functional correctness rates plateauing between 6.3\% and 13.7\%, which underscores the complexity of synthesizing valid operators for low-resource frameworks. The fast$_p$ metrics highlight the difficulty in achieving high-performance optimization. Even for the top-performing Claude-Sonnet-4.5, only 16.3\% of the generated operators match or exceed the baseline speed, with just 4.7\% realizing a significant speedup ($>1.5\times$). These findings substantiate the hypothesis that base large language models lack the intrinsic ability to discover hardware-efficient implementation strategies in the absence of external guidance or feedback.

\begin{figure*}[t]
\vspace{-8mm}
  \centering
  \includegraphics[width=\linewidth]{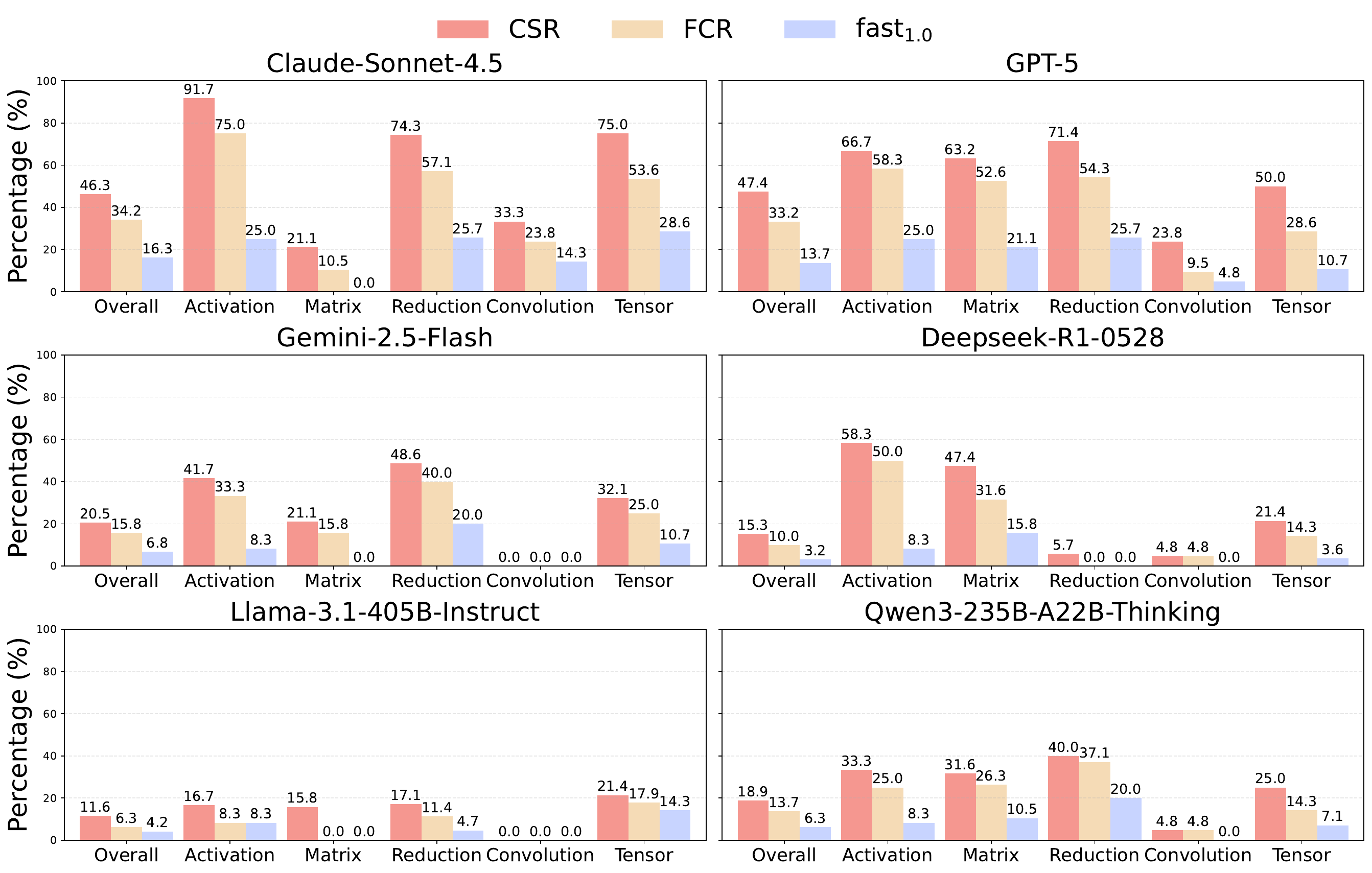}
  \caption{\textbf{Fine-grained performance of LLMs across different operator categories.} We visualize the evaluation metrics for five representative operator types. The results highlight significant disparities in model capabilities when handling operators with varying levels of algorithmic complexity.} 
  \vspace{-6mm}
  \label{fig:api_category}
\end{figure*}

We further dissect model performance across different operator categories, as illustrated in \cref{fig:api_category}. For computationally lightweight operations such as activation functions, most models exhibit strong performance, with Claude-Sonnet-4.5 achieving a functional correctness rate exceeding 70\%. In contrast, complex operators present a prohibitive challenge. Gemini-2.5-Flash and the leading open-source models fail to produce any functionally correct kernels for convolution tasks. Unexpectedly, GPT-5 and DeepSeek-R1-0528 display specific competence in matrix operations, achieving functional correctness rates of 52.6\% and 31.6\%, respectively. This observation suggests that these models may have retained effective algorithmic structures for general matrix multiplication during their pre-training phases, despite having lower overall performance averages.

\textbf{Supervised fine-tuning result by LoRA. } 
As shown in \cref{tab:main_results}, the base Qwen3-32B model demonstrates limited proficiency in this domain-specific task. Although SFT resulted in marginal improvements, specifically increasing the compilation success rate to 25.0\% and correctness to 18.5\%, it failed to yield any gains in the strict fast$_{1.5}$ efficiency metric. We attribute this limitation primarily to the inherent data scarcity characterizing mobile operator development. Given the constrained volume of available training samples, supervised learning proved insufficient for encoding the deep, framework-specific semantic knowledge required for MNN implementation. In contrast, our MoKA circumvents this bottleneck by actively retrieving optimization strategies via external tools. This distinction underscores that for low-resource, expert-domain tasks, an agentic framework capable of dynamic context retrieval and iterative reasoning constitutes a significantly more robust solution than static model fine-tuning.

\textbf{Reinforcement learning result by GRPO.}
As shown in \cref{tab:main_results}, the application of GRPO to the Qwen3-4B-Instruct-2507 model results in a significant 15.0\% improvement in the compilation success rate, effectively validating that our hierarchical reward design can successfully guide the model to comply with strict syntactic constraints. However, this method encounters limitations in terms of functional logic and performance optimization. The functional correctness rate only increases marginally by 2.5\%, and the high-performance metric fast$_{1.0}$ remains unchanged at 5.0\%. This stagnation indicates that although RL can align the model's output format, it fails to develop the complex reasoning skills needed to identify hardware-efficient strategies in smaller models. The superior performance of our MoKA demonstrates that iterative refinement is significantly more effective than training-time policy optimization for mobile operator generation.

\begin{table}[t]
  \centering
  \setlength{\tabcolsep}{8pt}
  \caption{\textbf{Comprehensive performance evaluation on MobileKernelBench.} We compare our proposed MoKA against standard prompting, SFT, and RL methods. CSR and FCR represent compilation success rate and functional correctness rate, respectively. Bold values indicate the best performance, and values in parentheses denote the absolute gains over the corresponding baselines.
  }
  \label{tab:main_results}
  \resizebox{\linewidth}{!}{
    \begin{tabular}{l c c c c c}
      \toprule
      \multirow{2}{*}{\textbf{Method}} & \textbf{CSR} & \textbf{FCR} & \multicolumn{3}{c}{\textbf{fast$_p$}} \\
      \cmidrule(lr){4-6}
       & (\%) & (\%) & \textbf{0.5} & \textbf{1.0} & \textbf{1.5} \\
      \midrule
      \multicolumn{6}{l}{\textit{\textbf{Supervised Fine-Tuning (SFT)}}} \\
      \addlinespace[1ex]
      Qwen3-32B & 16.7 & 13.9 & 8.3 & 2.8 & 0.0 \\
      Qwen3-32B-LoRA & 25 (+8.3) & 18.5 (+5.6) & 11.1 (+2.8) & 5.6 (+2.8) & 0.0 (--) \\
      \midrule
      \multicolumn{6}{l}{\textit{\textbf{Reinforcement Learning (RL)}}} \\
      \addlinespace[1ex]
      Qwen3-4B-Instruct-2507 & 10.0 & 5.0 & 5.0 & 5.0 & 0.0 \\
      Qwen3-4B-Instruct-2507-GRPO & 25.0 (+15.0) & 7.5 (+2.5) & 7.5 (+2.5) & 5.0 (--) & 0.0 (--) \\
      \midrule
      \multicolumn{6}{l}{\textit{\textbf{Agentic Workflow}}} \\
      \addlinespace[1ex]
      Claude-Sonnet-4.5 (Base) & 46.3 & 34.2 & 31.1 & 16.3 & 4.7 \\
      Claude-Sonnet-4.5 (@10) & 62.1 & 47.9 & 41.6 & 20.5 & 5.3 \\
      \rowcolor{gray!15} \textbf{MoKA (ours)} & \textbf{93.7} (+47.4) & \textbf{75.3} (+41.1) & \textbf{62.6} (+31.5) & \textbf{46.8} (+30.5) & \textbf{27.4} (+22.7) \\
      \bottomrule
    \end{tabular}
  }
\end{table}

\begin{figure*}[t]
  \centering
  \includegraphics[width=\linewidth]{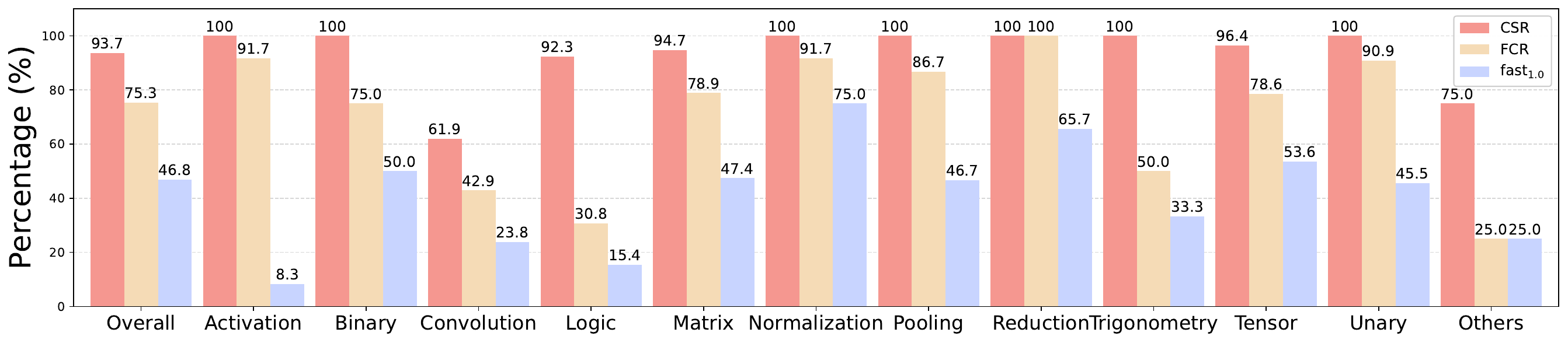}
  \caption{\textbf{Fine-grained performance of MoKA across different operator categories.}
  The agent demonstrates robust generalization across diverse operator types, achieving high correctness on structurally simpler operations (e.g., activation, normalization) while maintaining competitive performance on complex tasks like convolution and matrix operations.
  }
  \vspace{-6mm}
  \label{fig:agent_category}
\end{figure*}

\subsection{MoKA Results}
\label{sec:MoKA_result}
We initialize the MoKA using Claude-Sonnet-4.5 and perform N=10 iterations. To isolate the benefits of our agentic workflow from simple sampling diversity, we compare it against a pass@10 baseline where the same model is queried ten times.
As shown in \cref{tab:main_results}, while the pass@10 strategy yields a moderate improvement over baseline, the MoKA demonstrates superior performance across all metrics. Specifically, it achieves a functional correctness rate of 75.3\%, surpassing the single-query and pass@10 baselines by 41.1\% and 27.4\%, respectively. Most notably, in the speedup evaluation, the MoKA attains a 27.4\% success rate at the challenging fast$_{1.5}$ threshold, whereas baselines fail to exceed 6\%. This confirms that our iterative feedback loop enables the generation of not only correct but also highly efficient code.

\cref{fig:agent_category} further breaks down performance by operator category. The MoKA achieves a 100\% compilation success rate in seven categories and yields substantial correctness gains on complex tasks such as matrix ($>7\times$) and convolution (nearly $2\times$). These results indicate that the Debugger effectively leverages repository context to resolve hallucinations and dependency errors. Moreover, the average speedup of nearly $3\times$ across successful kernels validates the efficacy of the Accelerator in identifying hardware-specific optimizations.
\vspace{-2mm}
\section{Conclusion}
In this work, we investigate the automation of mobile kernel development using Large Language Models (LLMs). %
We introduce MobileKernelBench, a robust system coupled with an automated pipeline to holistically evaluate compilation success, functional correctness, and on-device performance. %
Our comprehensive evaluation of both off-the-shelf and fine-tuned models reveals that LLMs struggle with domain-specific mobile implementation, a failure we attribute to the knowledge deficits caused by ecosystem fragmentation and data scarcity. %
To bridge this gap, we propose MoKA, a framework that decomposes kernel refinement into cooperative agents for generation, debugging, and acceleration. %
MoKA achieves SOTA results, validating the efficacy of agentic optimization in this specialized domain. %
Our findings confirm that with principled methodological design, LLMs can effectively assist human developers in mobile kernel engineering. %
Future research directions include extending these capabilities to low-level optimizations such as multi-threaded NEON and inline assembly, exploring diverse hardware backends and frameworks beyond MNN, and potentially automating the generation of entire deployment frameworks.

\newpage
{
\small
\bibliographystyle{IEEEtran}
\bibliography{neurips_2026}

@String(NeurIPS = {Adv. Neural Inform. Process. Syst.})

@String(ICML  = {Int. Conf. Mach. Learn.})

@String(ICLR  = {Int. Conf. Learn. Represent.})

@String(NeurIPS = {NeurIPS})

@String(ICML  = {ICML})

@String(ICLR  = {ICLR})

@article{guo2024deepseek,
  title={DeepSeek-Coder: When the Large Language Model Meets Programming--The Rise of Code Intelligence},
  author={Guo, Daya and Zhu, Qihao and Yang, Dejian and Xie, Zhenda and Dong, Kai and Zhang, Wentao and Chen, Guanting and Bi, Xiao and Wu, Yu and Li, YK and others},
  journal={arXiv preprint arXiv:2401.14196},
  year={2024}
}

@article{hui2024qwen2,
  title={Qwen2. 5-coder technical report},
  author={Hui, Binyuan and Yang, Jian and Cui, Zeyu and Yang, Jiaxi and Liu, Dayiheng and Zhang, Lei and Liu, Tianyu and Zhang, Jiajun and Yu, Bowen and Lu, Keming and others},
  journal={arXiv preprint arXiv:2409.12186},
  year={2024}
}

@inproceedings{ouyang2025kernelbench,
  title={KernelBench: Can {LLM}s Write Efficient {GPU} Kernels?},
  author={Anne Ouyang and Simon Guo and Simran Arora and Alex L Zhang and William Hu and Christopher Re and Azalia Mirhoseini},
  booktitle={ICML},
  year={2025}
}

@article{wen2025multikernelbench,
  title={MultiKernelBench: A Multi-Platform Benchmark for Kernel Generation},
  author={Wen, Zhongzhen and Zhang, Yinghui and Li, Zhong and Liu, Zhongxin and Xie, Linna and Zhang, Tian},
  journal={arXiv eprints, pp. arXiv--2507},
  year={2025}
}

@inproceedings{li2025tritonbench,
  title={Tritonbench: Benchmarking large language model capabilities for generating triton operators},
  author={Li, Jianling and Li, Shangzhan and Gao, Zhenye and Shi, Qi and Li, Yuxuan and Wang, Zefan and Huang, Jiacheng and WangHaojie, WangHaojie and Wang, Jianrong and Han, Xu and others},
  booktitle={ACL},
  year={2025}
}

@article{baronio2025kevin,
  title={Kevin: Multi-turn rl for generating cuda kernels},
  author={Baronio, Carlo and Marsella, Pietro and Pan, Ben and Guo, Simon and Alberti, Silas},
  journal={arXiv preprint arXiv:2507.11948},
  year={2025}
}

@article{li2025autotriton,
  title={Autotriton: Automatic triton programming with reinforcement learning in llms},
  author={Li, Shangzhan and Wang, Zefan and He, Ye and Li, Yuxuan and Shi, Qi and Li, Jianling and Hu, Yonggang and Che, Wanxiang and Han, Xu and Liu, Zhiyuan and others},
  journal={arXiv preprint arXiv:2507.05687},
  year={2025}
}

@article{li2025cuda,
  title={Cuda-l1: Improving cuda optimization via contrastive reinforcement learning},
  author={Li, Xiaoya and Sun, Xiaofei and Wang, Albert and Li, Jiwei and Shum, Chris},
  journal={arXiv preprint arXiv:2507.14111},
  year={2025}
}

@inproceedings{wei2025astra,
  title={Astra: A Multi-Agent System for {GPU} Kernel Performance Optimization},
  author={Anjiang Wei and Tianran Sun and Yogesh Seenichamy and Hang Song and Anne Ouyang and Azalia Mirhoseini and Ke Wang and Alex Aiken},
  booktitle={NeurIPS 2025 Fourth Workshop on Deep Learning for Code},
  year={2025},
}

@article{zhang2025cudaforge,
  title={CudaForge: An Agent Framework with Hardware Feedback for CUDA Kernel Optimization},
  author={Zhang, Zijian and Wang, Rong and Li, Shiyang and Luo, Yuebo and Hong, Mingyi and Ding, Caiwen},
  journal={arXiv preprint arXiv:2511.01884},
  year={2025}
}

@misc{onnx,
  author = {Bai, Junjie and Lu, Fang and Ke, Zhang and others},
  title = {{ONNX}: Open Neural Network Exchange},
  year = {2019},
  publisher = {GitHub},
  journal = {GitHub repository},
  howpublished = {\url{https://github.com/onnx/onnx}},
  note = {Accessed: 2026-01-09}
}

@inproceedings{paszke2019pytorch,
  title={Pytorch: An imperative style, high-performance deep learning library},
  author={Paszke, Adam and Gross, Sam and Massa, Francisco and Lerer, Adam and Bradbury, James and Chanan, Gregory and Killeen, Trevor and Lin, Zeming and Gimelshein, Natalia and Antiga, Luca and others},
  booktitle={NeurIPS},
  year={2019}
}

@inproceedings{abadi2016tensorflow,
  title={TensorFlow: A System for Large-Scale Machine Learning},
  author={Abadi, Mart{\'\i}n and Barham, Paul and Chen, Jianmin and Chen, Zhifeng and Davis, Andy and Dean, Jeffrey and Devin, Matthieu and Ghemawat, Sanjay and Irving, Geoffrey and Isard, Michael and others},
  booktitle={OSDI},
  year={2016}
}

@inproceedings{chen2018tvm,
  title={TVM: An Automated End-to-End Optimizing Compiler for Deep Learning},
  author={Chen, Tianqi and Moreau, Thierry and Jiang, Ziheng and Zheng, Lianmin and Yan, Eddie and Shen, Haichen and Cowan, Meghan and Wang, Leyuan and Hu, Yuwei and Ceze, Luis and others},
  booktitle={OSDI},
  year={2018}
}

@inproceedings{jiang2020mnn,
  title={MNN: A universal and efficient inference engine},
  author={Jiang, Xiaotang and Wang, Huan and Chen, Yiliu and Wu, Ziqi and Wang, Lichuan and Zou, Bin and Yang, Yafeng and Cui, Zongyang and Cai, Yu and Yu, Tianhang and others},
  booktitle={MLSys},
  year={2020}
}

@software{Ni_ncnn_2017,
  author = {The ncnn contributors},
  license = {BSD-3-Clause},
  month = jun,
  title = {{ncnn}},
  url = {https://github.com/Tencent/ncnn},
  year = {2017}
}

@misc{opencl,
  author       = {{Khronos Group}},
  title        = {{OpenCL: Open Computing Language}},
  howpublished = {\url{https://www.khronos.org/opencl/}},
  year         = {2009},
  note         = {Accessed: 2026-01-09}
}

@misc{vulkan,
  author       = {{Khronos Group}},
  title        = {{Vulkan}},
  howpublished = {\url{https://www.khronos.org/vulkan}},
  year         = {2015},
  note         = {Accessed: 2026-01-09}
}

@misc{anthropic_claude_sonnet45,
  author       = {{Anthropic}},
  title        = {{Introducing Claude Sonnet 4.5}},
  howpublished = {\url{https://www.anthropic.com/news/claude-sonnet-4-5}},
  year         = {2025},
  month        = sep,
  note         = {Accessed: 2026-01-09}
}

@misc{openai_gpt5_2025,
  title        = {Introducing GPT-5},
  author       = {{OpenAI}},
  year         = {2025},
  month        = {Aug},
  howpublished = {\url{https://openai.com/index/introducing-gpt-5/}},
  note         = {Accessed: 2026-01-09}
}

@misc{deepmind_gemini,
  author       = {{Google DeepMind}},
  title        = {{Gemini Models – Next Generation AI Systems}},
  howpublished = {\url{https://www.deepmind.google/models/gemini/}},
  year         = {2025},
  note         = {Accessed: 2026-01-09}
}

@article{zeng2025glm,
  title={Glm-4.5: Agentic, reasoning, and coding (arc) foundation models},
  author={Zeng, Aohan and Lv, Xin and Zheng, Qinkai and Hou, Zhenyu and Chen, Bin and Xie, Chengxing and Wang, Cunxiang and Yin, Da and Zeng, Hao and Zhang, Jiajie and others},
  journal={arXiv preprint arXiv:2508.06471},
  year={2025}
}

@article{team2025kimi,
  title={Kimi k2: Open agentic intelligence},
  author={Team, Kimi and Bai, Yifan and Bao, Yiping and Chen, Guanduo and Chen, Jiahao and Chen, Ningxin and Chen, Ruijue and Chen, Yanru and Chen, Yuankun and Chen, Yutian and others},
  journal={arXiv preprint arXiv:2507.20534},
  year={2025}
}

@article{yang2025qwen3,
  title={Qwen3 technical report},
  author={Yang, An and Li, Anfeng and Yang, Baosong and Zhang, Beichen and Hui, Binyuan and Zheng, Bo and Yu, Bowen and Gao, Chang and Huang, Chengen and Lv, Chenxu and others},
  journal={arXiv preprint arXiv:2505.09388},
  year={2025}
}

@inproceedings{rajbhandari2020zero,
  title={Zero: Memory optimizations toward training trillion parameter models},
  author={Rajbhandari, Samyam and Rasley, Jeff and Ruwase, Olatunji and He, Yuxiong},
  booktitle={SC},
  year={2020},
}

@misc{qualcomm_snapdragon8gen2_productbrief_2022,
  author       = {{Qualcomm Technologies, Inc.}},
  title        = {{Snapdragon 8 Gen 2 Mobile Platform: Product Brief}},
  year         = {2022},
  howpublished = {PDF},
  publisher    = {{Qualcomm Technologies, Inc.}},
  url          = {https://www.qualcomm.com/content/dam/qcomm-martech/dm-assets/documents/Snapdragon-8-Gen-2-Product-Brief.pdf},
  urldate      = {2026-01-23},
  note         = {Product brief (2 pages)}
}

@software{tree-sitter2025,
  title = {tree-sitter/tree-sitter: An incremental parsing system for programming tools},
  author = {Applied Research Associates and GitHub and astral-sh and Disjunctive and sensmetry and AWS and University of Graz and Abacus AI and GuildEducationInc and Chime Systems Inc. and Intel-Corporation and River Point Technology and slashwhy},
  year = {2025},
  version = {v0.25.6},
  doi = {10.5281/zenodo.15594630},
  url = {https://github.com/tree-sitter/tree-sitter}
}

@article{shao2024deepseekmath,
  title={Deepseekmath: Pushing the limits of mathematical reasoning in open language models},
  author={Shao, Zhihong and Wang, Peiyi and Zhu, Qihao and Xu, Runxin and Song, Junxiao and Bi, Xiao and Zhang, Haowei and Zhang, Mingchuan and Li, YK and Wu, Yang and others},
  journal={arXiv preprint arXiv:2402.03300},
  year={2024}
}

@article{sheng2024hybridflow,
  title   = {HybridFlow: A Flexible and Efficient RLHF Framework},
  author  = {Guangming Sheng and Chi Zhang and Zilingfeng Ye and Xibin Wu and Wang Zhang and Ru Zhang and Yanghua Peng and Haibin Lin and Chuan Wu},
  year    = {2024},
  journal = {arXiv preprint arXiv: 2409.19256}
}

@article{su2025cuda,
  title={CUDA-L2: Surpassing cuBLAS Performance for Matrix Multiplication through Reinforcement Learning},
  author={Su, Songqiao and Sun, Xiaofei and Li, Xiaoya and Wang, Albert and Li, Jiwei and Shum, Chris},
  journal={arXiv preprint arXiv:2512.02551},
  year={2025}
}

@article{guo2025evoengineer,
  title={Evoengineer: Mastering automated cuda kernel code evolution with large language models},
  author={Guo, Ping and Zhu, Chenyu and Chen, Siyuan and Liu, Fei and Lin, Xi and Lu, Zhichao and Zhang, Qingfu},
  journal={arXiv preprint arXiv:2510.03760},
  year={2025}
}

@article{dubey2024llama,
  title={The llama 3 herd of models},
  author={Dubey, Abhimanyu and Jauhri, Abhinav and Pandey, Abhinav and Kadian, Abhishek and Al-Dahle, Ahmad and Letman, Aiesha and Mathur, Akhil and Schelten, Alan and Yang, Amy and Fan, Angela and others},
  journal={arXiv e-prints},
  year={2024}
}

@inproceedings{hu2022lora,
  title={Lora: Low-rank adaptation of large language models.},
  author={Hu, Edward J and Shen, Yelong and Wallis, Phillip and Allen-Zhu, Zeyuan and Li, Yuanzhi and Wang, Shean and Wang, Lu and Chen, Weizhu and others},
  booktitle={ICLR},
  year={2022}
}

@inproceedings{zheng2024llamafactory,
  title={LlamaFactory: Unified Efficient Fine-Tuning of 100+ Language Models},
  author={Yaowei Zheng and Richong Zhang and Junhao Zhang and Yanhan Ye and Zheyan Luo and Zhangchi Feng and Yongqiang Ma},
  booktitle={Proceedings of the 62nd Annual Meeting of the Association for Computational Linguistics (Volume 3: System Demonstrations)},
  year={2024},
}

@inproceedings{waghjale2024ecco,
  title={ECCO: Can we improve model-generated code efficiency without sacrificing functional correctness?},
  author={Waghjale, Siddhant and Veerendranath, Vishruth and Wang, Zhiruo and Fried, Daniel},
  booktitle={EMNLP},
  year={2024}
}

@inproceedings{huang2024effibench,
  title={Effibench: Benchmarking the efficiency of automatically generated code},
  author={Huang, Dong and Qing, Yuhao and Shang, Weiyi and Cui, Heming and Zhang, Jie M},
  booktitle={NeurIPS},
  year={2024}
}
}
\appendix
\section*{Overview}
\label{appdix:overview}
This appendix provides implementation details and experimental configurations to facilitate the reproducibility of MoKA. We begin by defining the taxonomy of MNN operators in \cref{appdix:MNN_OP}, which structures our code generation strategy based on operator implementation mechanisms. Building on this, \cref{appdix:prompts} presents the context-aware prompt templates designed for the Coder, Debugger, and Accelerator agents. To enable rigorous validation, \cref{appdix:pipeline} outlines the construction of our evaluation pipeline, covering repository restructuring, incremental cross-compilation, and on-device benchmarking protocols. We then elaborate on the GRPO training methodology in \cref{appdix:grpo_details}, detailing the compound reward formulation and the parallelized remote-mobile infrastructure used to handle sparse rewards and hardware constraints. Finally, \cref{appdix:case_study} provides a granular case study of the LayerNorm2D operator, illustrating the iterative optimization trajectory and the diverse hardware-aware strategies deployed by the agent to achieve significant speedups.

\section{MNN Operator Information}
\label{appdix:MNN_OP}
To ensure seamless integration with the MNN framework, we categorize all the operators into three distinct types based on their implementation mechanisms within MNN's architecture. This taxonomy aligns with the framework's operator lifecycle and dictates the specific source files targeted for code generation and replacement.
\begin{itemize}
    \item \textbf{Atomic operators:} This category encompasses fundamental operators that require direct, hardware-aware implementation on the CPU backend. These operators typically involve intensive numerical computation and cannot be trivially decomposed.
    \item \textbf{Geometric operators:} This category includes operators that can be mathematically expressed as coordinate transformations of input tensors.
    \item \textbf{Composite operators:} This category refers to high-level operators that do not have a direct one-to-one mapping in the MNN backend but can be represented as a composition of existing atomic operators.
\end{itemize}
Atomic operators require the synthesis of paired declaration (\texttt{.hpp}) and implementation (\texttt{.cpp}) files to define execution classes, while geometric and composite operators are encapsulated within a single source (\texttt{.cpp}) file. 
This structured classification enables our evaluation pipeline to automatically locate the corresponding C++ source files for injection and compilation, ensuring a robust and automated benchmarking process.

\section{Experiment Prompting Details}
\label{appdix:prompts}

\subsection{Initial Prompt for Coder}
\label{appdix:initial_prompt}
The Coder begins by querying the LLMs to generate C++ operator implementations. To bridge the gap between abstract PyTorch operator definitions and the architectural requirements of the MNN framework, we develop a dynamic, context-aware prompt generation mechanism. Instead of using a generic instruction, we construct a structured prompt tailored to the specific attributes of the target operator. Besides the PyTorch model definition of the target operator, the prompt also comprises the following key components:
\begin{itemize}
    \item \textbf{System role and task definition:} We define the LLM’s role as a mobile model deployment expert, with the explicit goal of converting a PyTorch model into a C++ operator for execution on the MNN CPU backend.
    \item \textbf{Constraint specification:} We enforce strict architectural constraints on the generated code, including adherence to internal APIs, implementation of required lifecycle methods (e.g., onResize, onExecute), and prescribed file naming and organization conventions.
    \item \textbf{Target operator context:} A key innovation of our pipeline is the dynamic injection of framework-specific knowledge. Based on the category of the target operator (e.g., unary, binary, reduction, or convolution), the relevant C++ header files are automatically retrieved and embedded into the prompt.
    \item \textbf{One-shot example:} To guide the code structure, we provide a paired example of PyTorch models and MNN C++ implementation, which aligns with the target implementation mechanism in the MNN library, corresponding to atomic, geometric, or composite operators.
\end{itemize}
To provide a concrete illustration of this context-aware generation mechanism, we present the full initial prompt used for the \texttt{ArgMax} operator below.
\begin{lstlisting}[language=prompt]
You are an expert in model deployment, proficient in PyTorch and C++ programming, and familiar with the coding style of the MNN framework. You will be given a PyTorch model which will be exported as an ONNX graph and then converted into an MNN computation graph. Your task is to write C++ code that implements and accelerates the operators from this model for MNN's CPU backend. Note that:
- Understand the example thoroughly and think carefully before you write the code.
- Provide only the code file as your final answer, without any explanations or comments.
- Your code must adhere to the supported API surfaces, invoking only official functions and members when using MNN C++ interfaces( e.g. Math, Tensor, VARP, Matrix) and flatbuffers library.
- Each file name must appear as the very first line inside its corresponding code block. If provide cpu backend implement, provide the hpp code and cpp code seperately. 
- Implement the operator's computation logic in a self-contained manner. Minimize coupling to MNN internals and 3rd party libraries; call APIs only when strictly required.
- Implement methods include: the CPU backend implement which handles numerical computation for operators by memory management and instruction-level optimization, the geometry computation which manages data layout and memory mapping and is used for operators that change tensor shapes or memory arrangements, the combinator implementation which builds new operator functions by composing existing MNN operators.
- When write CPU backend operator, implement onResize and onExecute. In onResize, allocate the cache buffer using backend()->onAcquireBuffer(&mCache, Backend::DYNAMIC) and release it with backend()->onReleaseBuffer(&mCache, Backend::DYNAMIC), allowing the freed memory to be reused. In onExecute, perform necessary input validation to catch issues early. Return NO_ERROR upon successful execution.
- When write a Geometry backend operator, implement onCompute to construct the tensor regions and command sequence that describe how outputs are assembled from inputs, allocating any intermediate tensors as virtual slices so the runtime can later schedule the actual computation efficiently.
- When write a Combiner operator, implement the onExecute method which parses the ONNX node's inputs and attributes to construct an equivalent computational subgraph using MNN's Express API and returns the converted expression to complete the operator translation.

Here is the example:
===== Example Start =====
Given PyTorch model Det:
[PyTorch code for Det model...]

By using CPU backend implemetation, you should respond with the final answer with two files seperately:

CPUDet.hpp:
``` 
#ifndef CPUDet_hpp
#define CPUDet_hpp
// ... [Includes and Class Definition]
class CPUDet : public Execution {
// ... [declarations for onResize, onExecute]
};
#endif
``` 

CPUDet.cpp:
``` 
#include "CPUDet.hpp"
// ... [Other Includes]

namespace MNN {
ErrorCode CPUDet::onResize(...) {
// ... [Buffer allocation logic]
return NO_ERROR;
}

ErrorCode CPUDet::onExecute(...) {
// ... [Complex determinant computation logic omitted]
return NO_ERROR;
}

class CPUDetCreator : public CPUBackend::Creator {
// ... [Creator implementation]
};
REGISTER_CPU_OP_CREATOR(CPUDetCreator, OpType_Det);
}
``` 
===== Example END =====

Now you are given the following PyTorch model:
``` 
import torch
import torch.nn as nn

class Model(nn.Module):
    """
    Simple model that performs Argmax over a specified dimension.
    """
    def __init__(self, dim: int):
        """
        Initializes the model with the dimension to perform argmax.

        Args:
            dim (int): The dimension to perform argmax over.
        """
        super(Model, self).__init__()
        self.dim = dim

    def forward(self, x: torch.Tensor) -> torch.Tensor:
        """
        Applies argmax over the specified dimension to the input tensor.

        Args:
            x (torch.Tensor): Input tensor.

        Returns:
            torch.Tensor: Output tensor with argmax applied, with the specified dimension removed.
        """
        return torch.argmax(x, dim=self.dim)

batch_size = 32
dim1 = 128
dim2 = 256

def get_inputs():
    x = torch.rand(batch_size, dim1, dim2)
    return [x]

def get_init_inputs():
    return [1]
```
Implement the CPU backend for the corresponding operator in this PyTorch model. You need to write the CPUArgMax.hpp and CPUArgMax.cpp files separately.
\end{lstlisting}

\subsection{Prompt for Debugger}
\label{appdix:debugger_prompt}
The Debugger is activated when the generated kernel code fails either during the compilation process or the functional correctness verification. We have designed two specialized prompt templates to address these distinct failure modes, ensuring the agent receives the relevant context needed to diagnose and resolve issues effectively.

When the MNN build system reports compilation errors, the constructed prompt includes:
\begin{itemize}
    \item \textbf{Operator context:} A description of the target operator, including its name, functionality, and the PyTorch reference model.
    \item \textbf{Current implementation:} The full source code generated by Coder that caused the failure.
    \item \textbf{Structured error log:} A parsed dictionary of compilation errors, categorized into \textit{in-place errors} (syntax or logic errors within the generated file) and \textit{cross-file errors} (mismatches with external MNN APIs or definitions). For each error, we provide the file path, line number, specific error message, and the relevant code context.
\end{itemize}
The agent is instructed to analyze these errors and provide semantic suggestions for code correction. Crucially, the instructions constrain the Debugger to suggest modifications \textit{only} for the current code fragment, treating external MNN files as immutable references. The output is required in a structured JSON format containing lists of suggestions for both in-place and cross-file errors.
Here is the prompt template and the provided error information:

\begin{lstlisting}[language=prompt]
'''
Here is a brief description of the operator to be implemented in MNN:
operator information:{op_info}

To implement this operator in MNN, the following script code has been implemented:
{code_book}

Compiling error accurred during MNN operator compilation, the information is listed as follows:
{compile_error}

Please analysize the compiling error and give suggestions to improve the MNN operator code.
Note:
 - Only provide semantic suggestions (in text) to improve the code.
 - Make sure your suggestion cover all the errors listed above.
 - Only provide semantic suggestions (in text) to improve the code.
 - All suggestions must be simple and direct.
 - Any suggestion involving code changes must modify only the current code snippet itself.
 - Code from other files is correct and for reference only. Do not suggest any modifications to other code.
 - For inplace errors, focus on correcting the code snippets provided in the error contexts.
 - For cross-file errors, refer to the relevant code snippets(function call, parameter settings, function defination etc.) in the error contexts and adjust the implementation currently provided code accordingly.
 - TODO: headers error.

 Finally provide suggestions as follows:
 ```error_suggestion
{
    {
        "local_error_suggestion":[],
        "crossfile_error_suggestion":[],
    }
}
 ```
'''
\end{lstlisting}

\begin{lstlisting}[language=prompt]

compile_error:{
"opname":"opname",
"local_error":{
    "erorr1":{
        "erorr_file":"path/to/error_file1",
        "error_line":123,
        "error_message":"Description of error 123",
        "error_line":456,
        "error_message":"Description of error 456",
        "error_context":"Context or code snippet related to error"
    }},
"crossfile_error":{
    "erorr1":{
        "erorr_file":"path/to/error_file1",
        "error_line":123,
        "error_message":"Description of error 123",
        "error_line":456,
        "error_message":"Description of error 456",
        "error_context":"Context or code snippet related to error 1" #use tree-sitter to extract code snippet(the whole fuction or class that include the error line)
    },
    "erorr2":""
},
"other_error":"other_error"
}
\end{lstlisting}

If the kernel compiles successfully but fails the correctness verification against the ONNX baseline, a different prompt is generated containing:
\begin{itemize}
    \item \textbf{Operator context \& implementation:} The same operator description and current code implementation as in the compilation error scenario.
    \item \textbf{Execution error log:} Detailed runtime error messages or output mismatch information captured during the test execution.
    \item \textbf{Model topology comparison:} JSON representations of both the reference ONNX model and the converted MNN model. This allows the agent to inspect graph-level discrepancies, such as incorrect attribute mapping or tensor shape mismatches.
\end{itemize}
The prompt directs the Debugger to compare the MNN and ONNX model structures alongside execution errors to identify logical flaws. Similar to the compilation error scenario, the agent must provide straightforward and actionable semantic suggestions for fixing the implementation script, formatted as a JSON list of items. Here is the template:

\begin{lstlisting}[language=prompt]
'''
Here is a brief description of the operator to be implemented in MNN:
operator information:{op_info}

To implement this operator in MNN, the following script code has been implemented:
{code_book}

The code has passed the compile process, but there are functionality correctness issues during testing. The execute information is listed as follows:
{execute_error}

The json files of onnx model and mnn model are:
onnx information:{onnx_json} \n
mnn information: {mnn_json} \n

Analysize the execution error and compare the information provided, then give suggestions to correct the MNN implemetation script code.
Note:
 - Only provide semantic suggestions (in text) to improve the code.
 - All suggestions must be simple and direct.
 - Any suggestion involving code changes must modify only the current code snippet itself.
 - Code from other files is correct and for reference only. Do not suggest any modifications to other code.
 - Refer to the differences between the MNN and ONNX json files, and propose reasonable code modification suggestions to acheive functionality correctness.
 
 Finally provide suggestions as follows:
 ```functionality_suggestion
{
    {
        "suggestion1":"",
        "suggestion2":"",
        # More ...
    }
}
 ```
'''
\end{lstlisting}

\subsection{Prompt for Accelerator}
\label{appdix:accelerator_prompt}
Once a generated kernel successfully passes both compilation and functional correctness verification, the Accelerator is engaged to further optimize its runtime performance. The prompt for this agent is designed to elicit a focused, high-impact optimization strategy rather than a broad list of potential improvements. The input context provided to the agent includes:
\begin{itemize}
    \item \textbf{Operator context \& implementation:} The same as Debugger.
    \item \textbf{Performance metrics:} The execution latency and relevant profiling data obtained from the on-device benchmarking of the current kernel.
    \item \textbf{Optimization history:} A record of previously attempted optimization strategies for this or similar operators. This is critical for preventing the agent from suggesting redundant or previously failed optimizations, ensuring efficient exploration of the solution space.
\end{itemize}
The instructions explicitly require the agent to identify exactly one primary performance bottleneck and propose exactly one corresponding optimization method expected to yield the largest speedup. Furthermore, the agent is directed to provide a concrete modification plan while keeping descriptions brief and technical. The output must be formatted as a JSON object containing three fields: bottleneck (diagnosis of the performance issue), optimization method (the proposed strategy), and modification plan (actionable steps for implementation). This structured output facilitates automated parsing and subsequent code generation cycles. Here is the prompt template:

\begin{lstlisting}[language=prompt]
'''
You are an expert in model deployment, proficient in PyTorch and C++ programming, and familiar with the coding style of the MNN framework. Your task is to analyse the performance bottlenecks of the following MNN operator code and propose optimisation methods to accelerate it.
Then identify **exactly one** highest-impact speed bottleneck, propose **exactly one** optimisation method and propose a modification plan.
Here is a brief description of the operator to be implemented in MNN:
operator information:{op_info}

To implement this operator in MNN, the following script code has been implemented:
{code_book}

Here is the current performance of the operator:
{performance}

Here is the history optimisation information of similar operators:
{history_optmz_info}

Requirements:
- Return **one and only one** optimisation method -- the largest expected speedup.
- Keep fields brief; avoid lists of alternatives, disclaimers, or generic advice.
- Avoid the totally same optimizations that have already been attempted in the history optimisation information.

Output format (JSON):
```json
{{
  "bottleneck": "<max 100 words>",
  "optimisation method": "<max 100 words>",
  "modification plan": "<max 100 words>"
}}
'''
\end{lstlisting}

\subsection{Iterative Refinement Prompt for Coder}
During the iterative optimization process, the Coder is re-engaged to modify the existing kernel implementation based on feedback from downstream agents. Depending on the state of the current kernel, the prompt is dynamically adjusted to focus either on code repair or performance optimization. The specific instructions diverge as follows:
\begin{itemize}
    \item \textbf{Repair mode:} When the kernel fails compilation or correctness tests, the prompt incorporates the Repair Suggestions generated by the Debugger agent. The Coder is explicitly tasked with refining the code to resolve these specific errors.
    \item \textbf{Acceleration mode:} When the kernel is functionally correct but requires speedup, the prompt incorporates the Optimization Plan generated by the Accelerator Agent. The Coder is tasked with implementing the proposed algorithmic or memory-level optimizations to improve execution latency.
    \end{itemize}
To ensure the output can be seamlessly integrated into the evaluation pipeline, strict formatting constraints are applied in both modes. The agent is required to generate only the C++ code without any accompanying natural language explanations or markdown commentary, which facilitates direct file overwriting and subsequent compilation cycles.

\section{Evaluation Pipeline Building from MNN}
\label{appdix:pipeline}
To instantiate our evaluation pipeline, we select MNN as the target framework for on-device inference. We begin by restructuring the MNN operator repository based on official documentation and internal operator semantics. As detailed in Appendix \ref{appdix:MNN_OP}, we dive into the codebase to decouple kernel implementations, ensuring that multiple kernels originally bundled in a single file are refactored into standalone files. This granular decoupling serves two purposes: it facilitates modular operator implementation and streamlines the registration process. The detailed configurations are as follows:

\textbf{Operator registration}. Given that MNN's support for PyTorch operators is still evolving, we utilize ONNX as the intermediate representation for evaluation, adhering to official ONNX operator design principles. Our registration mechanism employs a semantic matching strategy, mapping operator names to their corresponding registration schemas and kernel directory paths within the source tree. During experimentation, we implement a ``protected hot-swapping" mechanism: a backup of the original operator implementation is created before replacing it with the generated kernel code. Once the cycle of compilation, verification, and benchmarking is complete, the original implementation is restored, preserving the integrity of the operator repository.

\textbf{Incremental compilation.} We establish our build environment on \textbf{Ubuntu 24.04.3 via WSL} using the GCC toolchain. To optimize efficiency, we first pre-compile the entire framework foundation. Following operator registration, we employ an incremental compilation strategy to rebuild the framework and generate the necessary model conversion (\textbf{MNNConvert}) and benchmarking tools. For mobile deployment, we utilize the \textbf{Android NDK toolchain} to cross-compile the adapted MNN library and binaries for the Android runtime environment.

\textbf{Correctness verification.} To validate functional correctness, we leverage MNN's native model conversion tools. The tool automatically computes the discrepancy between the outputs of the source ONNX model and the converted MNN model under identical input conditions. We enforce a strict numerical tolerance threshold of 1e-4.

\textbf{On-Device performance benchmarking.} Upon passing verification, we conduct performance profiling on the target mobile device. We first prepare the cross-compiled MNN benchmarking binary and the modified model on the host system. These assets are deployed to the device's /data/local/tmp directory via the \textbf{Android Debug Bridge (ADB)}. To ensure statistical robustness, we adopt a multi-iteration strategy, executing the inference loop \textbf{100 times} to calculate the average latency per operator. Throughout the testing phase, we monitor execution to ensure the device remains in a quiescent state, maintaining a CPU utilization \textbf{below 10\%} (relative to total capacity of 800\% on an 8-core system) to minimize system noise and thermal throttling. Performance metrics are analyzed using MNN's profiling output\footnote{https://mnn-docs.readthedocs.io/en/latest/tools/test.html}.

This evaluation configuration aims to provide a fair and stable evaluation environment where all operators are measured under consistent device conditions. In real-world mobile workloads, operator execution typically occurs alongside other system activities and model components, where resource contention, memory bandwidth pressure, and thermal effects may reduce the achievable performance. However, accurately reproducing such dynamic system loads in a controlled and reproducible manner is challenging. Similar to common practices in GPU kernel benchmarking, we therefore isolate operator execution and minimize background utilization to approximate the intrinsic computational performance of each operator. In practice, a baseline CPU utilization of around 10\% corresponds to the typical system overhead observed on an idle device after boot, providing a reasonable compromise between realism and measurement stability.

\section{GRPO Training Details}
\label{appdix:grpo_details}
In this section, we detail the implementation strategies used during the GRPO training, focusing on reward design, parallel evaluation infrastructure, and cross-platform hardware connectivity.

\textbf{Reward design.}
We base our reward formulation on the structure proposed in previous work cite{baronio2025kevin}, which incentivizes both correctness and latency reduction. The original baseline reward function is defined as:
\begin{equation}
    \text{Reward}= 0.3 \cdot 1_{\{\text{correct}\}} + \frac{T_{\text{baseline}}}{T_{\text{generated}}} \cdot 1_{\{\text{correct}\}}
\end{equation}
where $1_{\{\text{correct}\}}$ is an indicator that equals 1 if the generated kernel passes numerical verification against the ONNX baseline, and 0 otherwise. $T_{\text{baseline}}$ and $T_{\text{generated}}$ represent the inference latency of the baseline and generated kernels, respectively.
However, generating C++ kernels for the MNN mobile framework presents distinct challenges compared to CUDA kernel generation. The strict reliance on MNN's internal APIs and memory lifecycle often leads to code that fails to compile initially. Consequently, the model faces the ``sparse reward'' problem, where the reward remains zero for extended periods, hindering the learning of syntactic and structural correctness. To mitigate this, we introduce an intermediate compilation reward. Our modified reward signal is defined as:
\begin{equation}
    \text{Reward}= 0.3 \cdot 1_{\{\text{compile}\}} + 0.3 \cdot 1_{\{\text{correct}\}} + \frac{T_{\text{baseline}}}{T_{\text{generated}}} \cdot 1_{\{\text{correct}\}}
\end{equation}
By adding the term $0.3 \cdot 1_{\{\text{compile}\}}$, we provide the model with early feedback on syntactical validity, effectively guiding the policy optimization process through the initial cold-start phase.

\textbf{Parallelized evaluation environment.}
During the GRPO training phase, the policy generates a group of 5 code samples for each prompt. Since the MNN build system involves complex file dependencies and intermediate object generation, multiple compilation processes cannot share the same working directory without causing race conditions or linkage errors.
To enable efficient parallel evaluation, we implement a workspace isolation strategy. For each generated sample in a group, we instantiate an independent working directory by duplicating the necessary MNN core library and build scripts. This allows the compilation and correctness verification of multiple samples to proceed concurrently, significantly accelerating the reward computation without mutual interference.

\textbf{Remote-mobile bridge connection.}
Since our training pipeline runs on a remote high-performance Linux server cluster, while the performance measurements must be executed on real-world mobile devices, we established a seamless bridge between the two environments.
We use secure shell (SSH) reverse tunneling to forward the local ADB server socket from the host machine to the remote Linux server. This setup allows the training pipeline on the remote server to issue commands such as \texttt{adb push} and \texttt{adb shell} directly on the mobile device connected to the local host. This architecture supports full bidirectional file transfer and shell execution, enabling the evaluation script to deploy compiled shared libraries to the mobile device and retrieve profiling logs automatically.

\section{Case Study}
\label{appdix:case_study}

\begin{figure*}[t]
  \centering
  \includegraphics[width=\linewidth]{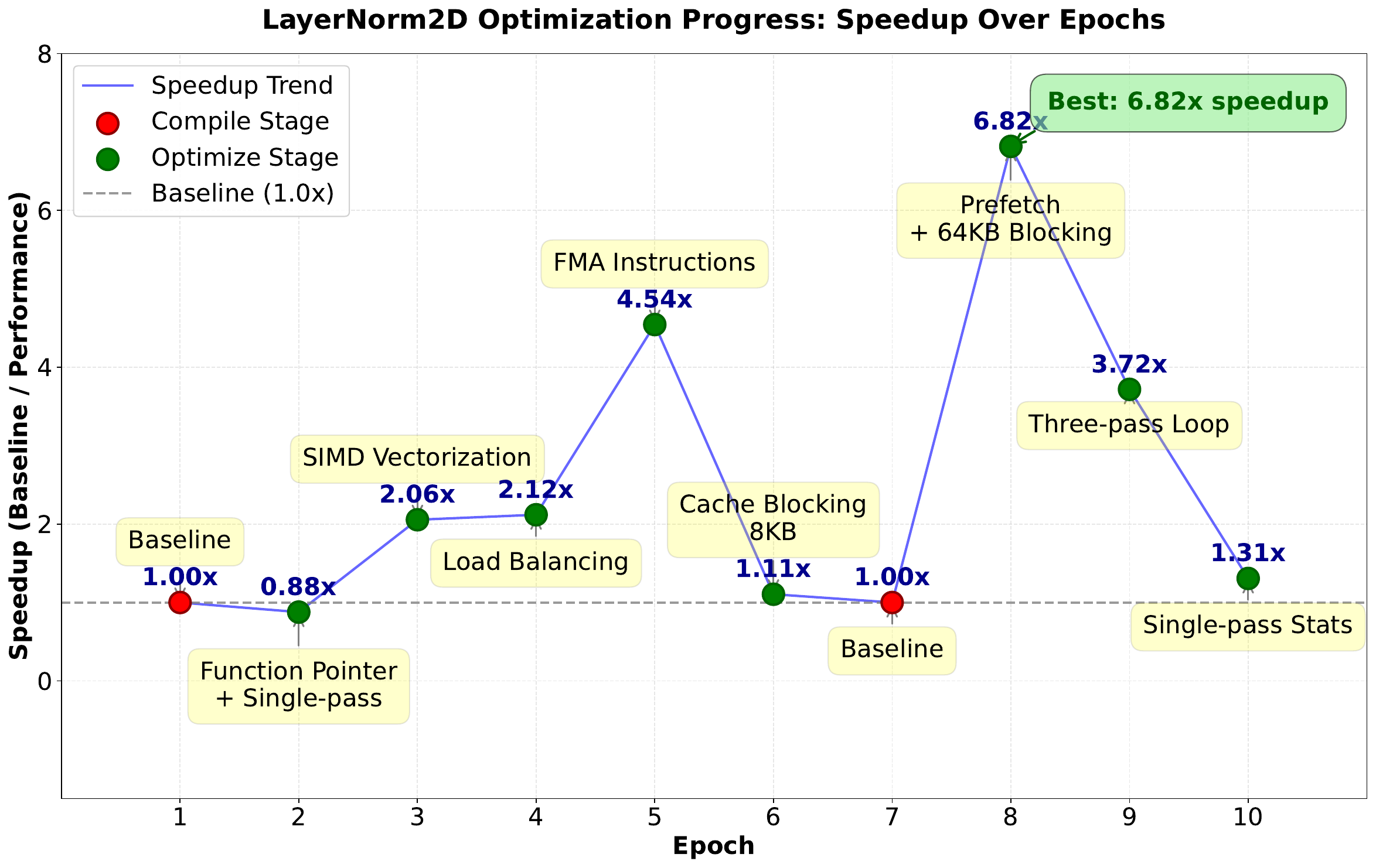}
  \caption{\textbf{MoKA optimize LayerNorm2D kernel process.} We conduct the optimization process for 10 iterations. MoKA shows that LLMs can provide diverse optimization methods like SIMD and cache blocking, achieve remarkable speedups.} 
  \label{fig:case_study}
\end{figure*}


\newcolumntype{Y}{>{\centering\arraybackslash}X} 

\begin{table}[t]
\centering
\caption{\textbf{Case study: LayerNorm2D.} We demonstrate MoKA's capability to automatically optimize kernel implementations through 10 iterative epochs. MoKA achieves a peak speedup of 6.82× at epoch 8, perform an average speedup of 2.82x. }

\label{tab:case_study}

\def\TableWidth{\textwidth}
\renewcommand{\arraystretch}{1.1} 
\small

\begin{tabularx}{\TableWidth}{l Y Y Y  Y } 
\toprule
\textbf{Epoch} & \textbf{Stage} & \textbf{Performance / ms} & \textbf{Baseline / ms}& \textbf{Speedup}\\
\midrule
    1 & compilation & -- & 0.409  & -- \\
    2 & optimization  & 0..466  & 0.409  & 0.88 \\
    3 & optimization  & 0.199  & 0.409  & 2.06\\
    4 & optimization  & 0.193  & 0.409  & 2.12\\
    5 & optimization  & 0.09  & 0.409 & 4.54\\
    6 & optimization  & 0.37  & 0.409 & 1.11\\
    7 & compilation  & --  & 0.409 & -- \\
\rowcolor{gray!15}    8 & optimization  & \textbf{0.06}  & 0.409 & \textbf{6.82}\\
    9 & optimization  & 0.11  & 0.409 & 3.72\\
    10 & optimization  & 0.313  & 0.409 & 1.31\\
\bottomrule
\end{tabularx}
\end{table}

As shown in \cref{tab:case_study} and \cref{fig:case_study}, starting from a baseline implementation (1.00x), MoKA demonstrates remarkable efficacy in navigating the complex optimization space, ultimately achieving a peak speedup of 6.82x at epoch 8. The process reveals the agent's ability to systematically identify and address hierarchical bottlenecks. Initially, MoKA targets computational efficiency, introducing SIMD vectorization (epoch 3, 2.06x) and FMA instructions (epoch 5, 4.54x) to maximize instruction throughput. As the compute bound is alleviated, the agent autonomously shifts focus to memory latency. By epoch 8, MoKA successfully implements advanced memory-hiding techniques—specifically 64KB cache blocking with software prefetching—to overcome the memory wall, yielding the global optimal performance. This trajectory confirms that MoKA can effectively synthesize diverse optimization strategies, ranging from instruction-level parallelism to memory hierarchy management, without human intervention.

\textbf{Performance fluctuations and optimization path diversity.} It is worth noting that the optimization curve exhibits a non-monotonic "sawtooth" pattern, particularly the significant performance drops observed at epoch 6 and epoch 9. This phenomenon stems from our design constraint where the LLM is prompted to identify and resolve only the single most critical bottleneck in each iteration. This focus forces the agent to choose specific \textit{optimization lanes} that may conflict with previous gains. For instance, the transition from epoch 5 to epoch 6 involved \textbf{a shift from a compute-centric strategy (FMA) to a memory-centric strategy (small-block Welford algorithm)}. The overhead introduced by the complex memory access pattern in epoch 6 inadvertently negated the computational gains, causing a temporary regression. However, this volatility is instrumental; the performance drop serves as negative feedback, prompting the agent to self-correct. In subsequent iterations, MoKA refined the blocking strategy and reintroduced prefetching, successfully reconciling the conflict between compute and memory optimizations to reach the global optimum.




\end{document}